%% file: iclr2027_conference.tex
\documentclass{article} 
\usepackage{iclr2027_conference,times}

\input{math_commands.tex}

\usepackage{hyperref}
\usepackage{url}
\usepackage{graphicx}
\usepackage{float}
\usepackage{booktabs}
\usepackage{multirow}
\usepackage{adjustbox}
\usepackage{array}
\usepackage{comment}

\title{When to Trust Imagination: Adaptive Action Execution for World Action Models}

\author{%
\makebox[\dimexpr\textwidth-2\tabcolsep\relax][c]{%
\normalfont
\begin{tabular}[t]{@{}c@{}}
\rule{0pt}{1.0em}%
\textbf{Rui Wang}$^{1,3}$
\quad
\textbf{Yue Zhang}$^{2}$
\quad
\textbf{Canyang Chen}$^{1}$
\quad
\textbf{Jiehong Lin}$^{2}$
\\[0.3em]
\textbf{Zhongrui Wang}$^{1}$\thanks{
Corresponding authors:
Xiaojuan Qi (\texttt{xjqi@eee.hku.hk}) and
Zhongrui Wang (\texttt{wangzr@sustech.edu.cn}).
}
\quad
\textbf{Xiaojuan Qi}$^{2,3}$\footnotemark[1]
\\[0.6em]
$^{1}$Southern University of Science and Technology
\quad
$^{2}$The University of Hong Kong
\\
$^{3}$Shenzhen Loop Area Institute
\end{tabular}%
}%
}

\iclrfinalcopy
\begin{document}

\maketitle
\lhead{}

\begin{abstract}

World Action Models (WAMs) jointly predict future visual observations and actions, but typically execute a fixed number of actions before replanning, regardless of whether the imagined future remains consistent with reality. We formulate adaptive WAM execution as \emph{future--reality verification} and propose \emph{Future Forward Dynamics Causal Attention} (FFDC), a lightweight verifier that reasons over predicted actions, predicted visual dynamics, real observations, and instruction-conditioned semantics. FFDC allows continued execution when a plan remains reliable and triggers replanning when its validity deteriorates. We combine verification with Mixture-of-Horizon Training to improve long-horizon trajectory coverage. On RoboTwin, FFDC-WAM achieves the highest average success rate among the evaluated methods in both clean and randomized settings. Under randomization, it improves success by 2.54 percentage points over Base-Motus while reducing WAM planning calls by 69.10\% and completion time by 33.88\%.  Notably, across six real-world tasks on a Unitree G1D robot, FFDC-WAM improves average success by 9.17\% over Base-Motus-LC16 while reducing calls by 74.06\% and completion time by 51.31\%.

\end{abstract}

\section{Introduction}
\label{sec:introduction}

Humans anticipate the consequences of their actions and use sensory feedback to decide whether to continue. When a foot misses an expected stair, the unexpected lack of contact prompts an immediate correction; when events unfold as expected, we can proceed without reconsidering every movement. This comparison between prediction and observation allows efficient action while preserving the ability to respond to change.

World Action Models (WAMs) offer a computational analogue. Whereas many vision-language-action policies generate actions from observations and instructions~\citep{shukor2025smolvla, black2024pi_0, black2025pi_, lv2025f1}, WAMs jointly predict future actions and visual observations~\citep{pai2025mimic, liang2025video, li2026causal, du2023learning, ye2024latent, guo2024prediction}. Recent WAMs have shown promising transfer across tasks, environments, and embodiments~\citep{bi2025motus, li2025unified, zhu2025unified, ye2026gigaworld, ye2026world, yuan2026fast, beingh07}. Yet their predicted futures are seldom used to decide \emph{when} to replan during execution. Instead, the robot typically executes a fixed number of predicted actions before calling the WAM again.

\begin{figure}[h]
    \centering
    \includegraphics[width=\linewidth, trim={20 170 20 20}, clip]{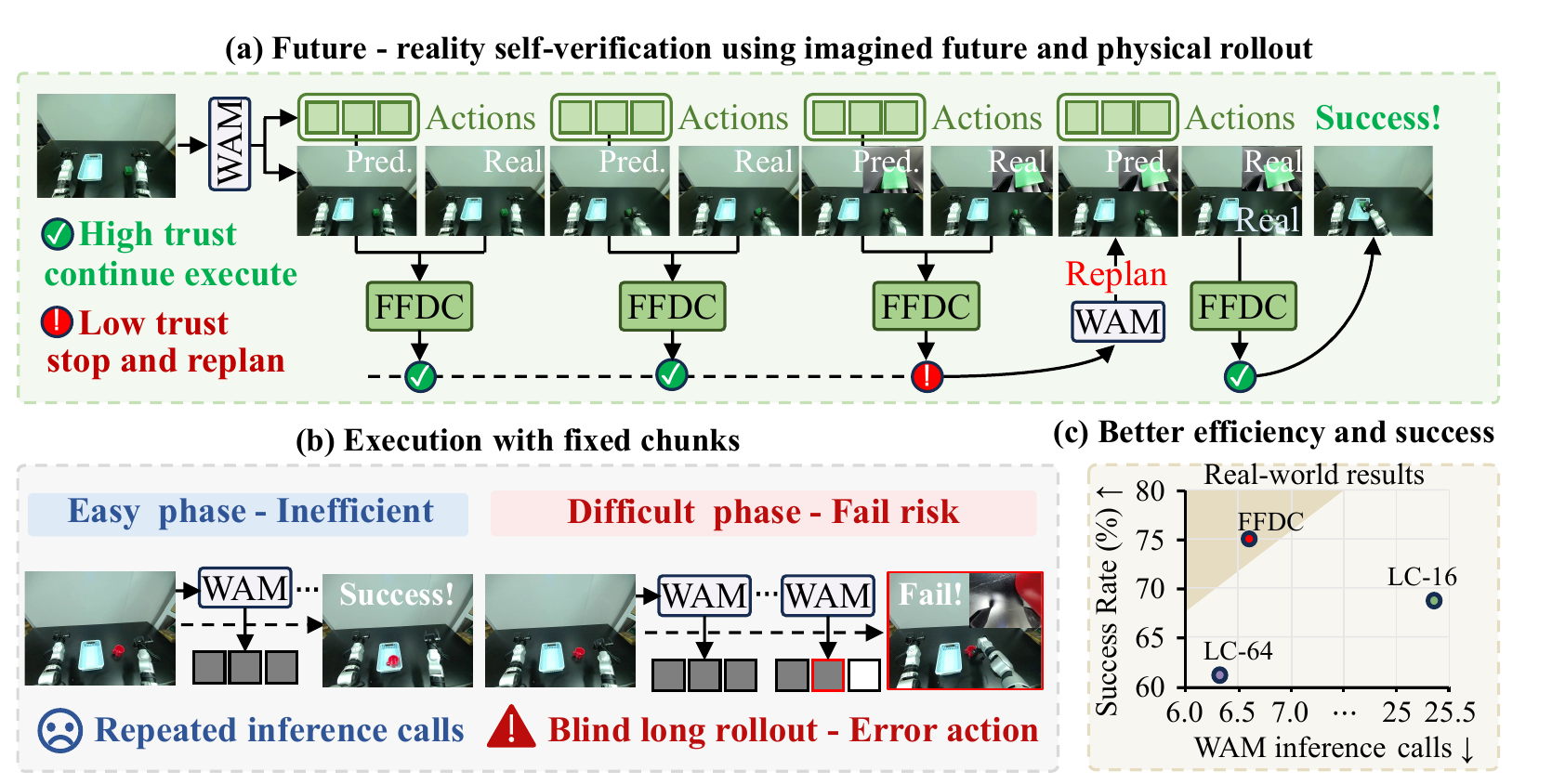}
    \caption{FFDC enables adaptive trust in WAM imagination. (a) FFDC verifies future--reality consistency to continue execution or trigger replanning. (b) Fixed execution horizons are inefficient in predictable phases and unreliable in difficult ones. (c) Success rate versus WAM calls on real-world tasks, illustrating FFDC's robustness--efficiency trade-off.}
    \label{fig:fig1}
\end{figure}

A fixed execution horizon creates an accuracy--latency tension (see Fig. \ref{fig:fig1} (c)). When the world evolves predictably, a plan may remain useful for many steps; repeatedly invoking WAM after short action chunks wastes time. During dynamic or contact-rich interaction, unexpected object motion or changes in contact can invalidate the remaining actions. Real-world actuation variability and noise increase the risk of such deviations. Short chunks provide frequent opportunities to correct the plan, but incur the cost of frequent WAM inference. The appropriate execution length therefore varies not only across tasks but also across phases of the same task. Therefore, the key challenge is not merely choosing a better chunk size, but deciding \emph{when the WAM's imagined future should still be trusted during physical execution}.

Existing adaptive execution methods adjust chunk length using action uncertainty, entropy, or policy-side confidence~\citep{jing2025mixture, wang2026vla, liang2026adaptive, wang2026open, wu2026speedup}. These signals reflect confidence in predicted actions, but may not reveal whether those actions have changed the world as expected. Unexpected object motion or contact can invalidate even a confident plan. WAMs offer complementary evidence: their predicted visual dynamics provide an expected rollout against which new observations can be assessed without another WAM call. Yet visual mismatch alone is an unreliable trigger. Benign deviations may leave the plan executable, while a subtle change in pose or contact may invalidate the next action. A verifier must therefore consider the latest observation together with the predicted visual trajectory, remaining actions, and task instruction to decide whether to continue or replan.

Based on this insight, we propose an adaptive WAM execution framework that explicitly compares WAM imagination with physical rollout. The core module is \emph{Future Forward Dynamics Causal Attention} (FFDC), a lightweight verifier that estimates whether the remaining WAM-predicted action segment is still reliable. As shown in Figure~\ref{fig:fig1} (a), the WAM predicts future visual dynamics and an action chunk, while FFDC verifies during execution whether the imagined future remains consistent with the real observation, planned actions, and language instruction. FFDC uses a structured attention mechanism to model the interaction between predicted vision and action, allowing it to detect task-critical mismatches and decide whether the remaining rollout can still be trusted. To equip FFDC with the ability to distinguish reliable imagined futures from deviations that require replanning, we construct a binary verification dataset using valid segments from demonstrations and successful rollouts, together with failure-prone segments from failed rollouts and synthetic action corruptions, and train it to predict the executability of the remaining action segment.

FFDC turns fixed-horizon WAM rollout into adaptive execution, continuing while new observations support the remaining actions and replanning when they do not. It thus executes longer in predictable phases and corrects sooner when the plan becomes unreliable. In real-world experiments, FFDC achieves the highest success rate among the evaluated methods with far fewer WAM calls than fixed short-chunk execution (Figure~\ref{fig:fig1}(c)). {Notably, across six real-world manipulation tasks, FFDC improves average success by 13.33\% over fixed 64-action execution with only 2.33 seconds of additional completion time. Compared with fixed 16-action execution, it reduces WAM calls by 74.06\% and completion time by 51.31\%, showing that selective replanning improves robustness under real-world variability without frequent WAM inference.}

Our contributions are summarized as follows:
\begin{itemize}
    \item We formulate adaptive WAM execution as a \emph{future--reality verification} problem, where the WAM's predicted visual future is used to assess whether its remaining action rollout can still be trusted.
    \item We propose \emph{Future Forward Dynamics Causal Attention} (FFDC), a lightweight verifier that models temporally aligned causal interactions between predicted actions, predicted visual dynamics, real observations, and language instructions to detect unreliable future execution.
    \item We show that FFDC enables adaptive trust in WAM imagination: long execution in predictable phases to reduce inference cost, and short execution in difficult phases to improve robustness. Experiments on RoboTwin and in the real world validate our method.
\end{itemize}

\section{Related work}

\paragraph{World action models.}
\label{sec:world action models}
World Action Models (WAMs) extend VLA policies through joint video-action modeling, coupling action generation with future visual dynamics~\citep{zhu2025unified,li2025unified,bi2025motus,ye2026world,li2026causal,yuan2026fast}. Unified formulations support forward dynamics, inverse dynamics, and conditional action or video generation, providing rich supervision for learning physical interactions~\citep{zhu2025unified,bi2025motus,li2025unified}. However, predicted futures are often used primarily for representation learning or action generation. To reduce inference cost, some methods rely on latent world features or omit explicit future rollout at test time~\citep{yuan2026fast,ye2026gigaworld}. This leaves the use of predicted futures for online execution verification underexplored. Our work uses WAM predictions as expectations of physical rollout, comparing them with real observations to determine whether the remaining action sequence can still be trusted.

\paragraph{Adaptive action execution.}
\label{sec:adaptive action execution}
Adaptive execution addresses the limitations of fixed open-loop rollout. Early approaches mitigate compounding errors through corrective supervision and recovery training~\citep{ross2011reduction,laskey2017dart}, while subsequent methods use uncertainty or risk estimates to trigger intervention or replanning~\citep{menda2019ensembledagger,hoque2021thriftydagger,lee2025diff}. Recent diffusion-policy and VLA methods adapt execution through multi-horizon prediction~\citep{jing2025mixture}, entropy-based chunk selection~\citep{liang2026adaptive}, verification~\citep{wang2026open}, learned scheduling~\citep{wu2026speedup}, or executable-horizon estimation~\citep{wang2026vla}. Self-Guidance and Adaptive Chunking (SGAC)~\citep{so2026improving}, our directly comparable adaptive baseline, retains or replaces the action queue based on similarity between previously planned and newly predicted actions, requiring fresh policy inference for consistency checks. FFDC instead exploits cached WAM predictions to perform lightweight \emph{future--reality verification}, jointly reasoning over predicted visual dynamics, real observations, planned actions, and task semantics to decide whether to continue execution or replan.

\section{Method}
\label{sec:method}
In this section, we present FFDC-WAM, a framework that combines low-frequency macro planning with high-frequency lightweight verification for efficient adaptive action execution by leveraging the joint video-action modeling capability of WAMs. We first introduce the standard action-chunking method in WAMs and adaptive action execution in Section~\ref{sec:preliminary}. We then present the architecture of FFDC-WAM in Section~\ref{sec:ffdc}, where a lightweight verifier performs high-frequency verification through a causal attention mechanism over visual and action modalities. Finally, in Section~\ref{sec:training strategy}, we describe the training strategies for the long-horizon WAM and the verifier module.

\subsection{Preliminary}
\label{sec:preliminary}
\paragraph{World action model with action chunking.}

We build on Motus~\citep{bi2025motus}, a world action model (WAM) that jointly predicts future actions and future visual observations conditioned on the current observation and language instruction. During training, the model is optimized with rectified flow-matching losses for both action and video prediction:
\begin{equation}
\mathcal{L}_{\text{WAM}} = \mathcal{L}_{\text{act}} + \mathcal{L}_{\text{vid}}.
\end{equation}

At inference time, given the current observation \(o_t\) and instruction \(\ell\), the WAM predicts a future action chunk and corresponding latent future visual tokens:
\begin{equation}
\label{eq:wampred}
(\hat{A}_{t+1:t+H}, \hat{O}_{t+1:t+H}) = \pi_\theta(o_t, \ell),
\end{equation}
where \(\hat{A}_{t+1:t+H}\) denotes the predicted action chunk of length \(H\), and \(\hat{O}_{t+1:t+H}\) denotes the predicted latent visual sequence.

\paragraph{Adaptive action execution.}
Standard action chunking executes the predicted chunk in an open-loop manner and replans only after all \(H\) actions are finished. While efficient, this fixed execution scheme can accumulate errors in dynamic or contact-rich scenarios.

To enable adaptive execution, we introduce a verifier \(\mu_\phi\) that decides whether the remaining predicted rollout is still trustworthy. After executing part of the current chunk, the verifier takes the latest observation, predicted future actions, predicted future visual tokens, and instruction as input:
\begin{equation}
e_t =
\mu_\phi\!\left(
o_t,\hat{A}_{t:t+k},\hat{O}_{t:t+k},\ell
\right),
\end{equation}
where \(e_t \in [0,1]\) is a confidence score. The robot continues execution if \(e_t \ge \tau\), and replans otherwise:
\begin{equation}
\label{eq:execution_decision}
\text{execute if } e_t \ge \tau,\qquad
\text{replan if } e_t < \tau,
\end{equation}
where \(\tau=0.5\) in this paper. The objective of adaptive execution is to retain the efficiency advantage of chunked inference while restoring responsiveness to execution failures and environmental changes.

\subsection{Future forward dynamics causal attention}
\label{sec:ffdc}

\begin{figure}[t]
  \centering
  \includegraphics[width=1\linewidth, height=1\textheight, keepaspectratio,
                   trim=0 274 0 245, clip]{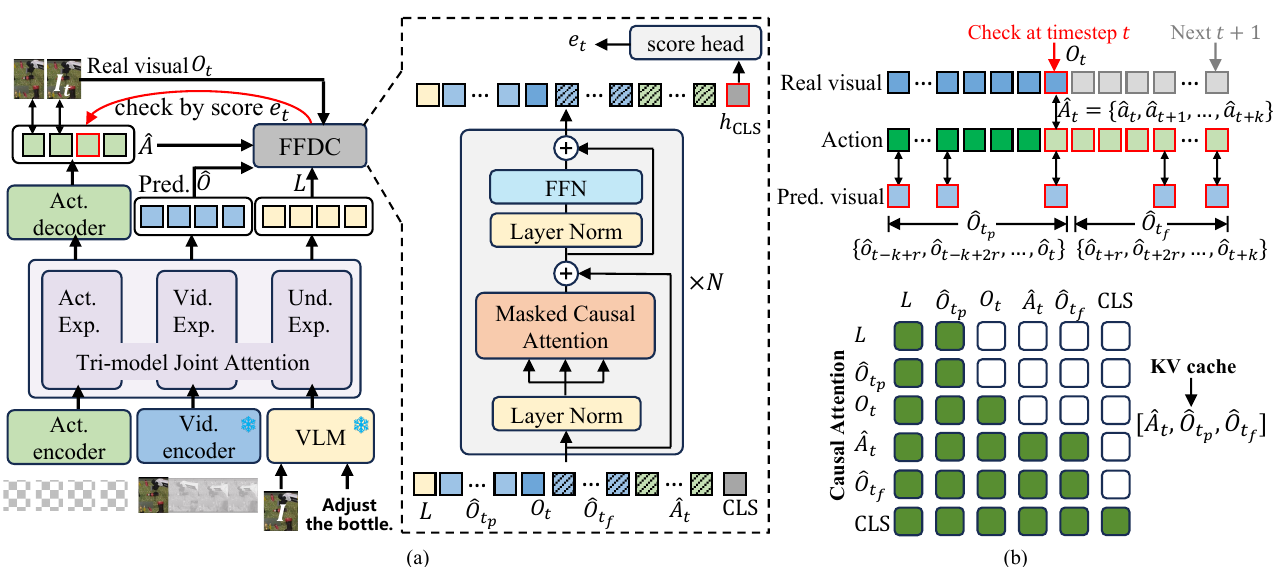}
  \caption{Overview of the proposed FFDC-WAM. (a) Given the action sequence, predicted video tokens, and semantic tokens generated by WAM, the FFDC verifier outputs an execution confidence score for the remaining plan. (b) At each check step \(t\), FFDC performs structured causal attention, enforcing temporally aligned interaction between action and predicted visual dynamics.}
  \label{fig:fig2}
\end{figure}

\paragraph{Verifier architecture.}

To determine whether the remaining predicted plan is still reliable under the latest observation, we introduce a verifier based on \emph{Future Forward Dynamics Causal Attention} (FFDC). As illustrated in Figure~\ref{fig:fig2} (a), each WAM inference produces a predicted action sequence \(\hat{A}\), the corresponding latent video tokens \(\hat{O}\), and the semantic tokens \(L\) from the Understanding expert.

At verification step \(t\), the verifier takes as input the current real observation tokens \(O_t\), the semantic tokens \(L\), the WAM-predicted historical video tokens \(\hat{O}_{t_p}\), the WAM-predicted future video tokens \(\hat{O}_{t_f}\), the future action segment \(\hat{A}_t\), and a learnable \texttt{[CLS]} token for global aggregation. The resulting verifier input sequence is
\begin{equation}
X_t = [L,\hat{O}_{t_p},O_t,\hat{O}_{t_f},\hat{A}_t,\texttt{[CLS]}].
\end{equation}

\paragraph{Future forward dynamics causal attention.}
To verify whether the next predicted action segment remains executable, we consider a horizon-\(k\) candidate rollout
\(
\hat{A}_t=[\hat{a}_t,\hat{a}_{t+1},\dots,\hat{a}_{t+k}]
\).
We also collect temporally aligned WAM-predicted visual tokens around timestep \(t\), including a past segment
\(
\hat{O}_{t_p}=[\hat{o}_{t-k+r},\hat{o}_{t-k+2r},\dots,\hat{o}_{t}]
\)
and a future segment
\(
\hat{O}_{t_f}=[\hat{o}_{t+r},\hat{o}_{t+2r},\dots,\hat{o}_{t+k}]
\),
where \(r\) is the action-to-video frequency ratio. In addition, we use instruction-conditioned semantic tokens \(L\) from the understanding expert and the latest real observation token \(O_t\).

A key design choice is that the WAM-predicted tokens, including past/future visual tokens \((\hat{O}_{t_p}, \hat{O}_{t_f})\), action tokens \(\hat{A}_t\), and understanding-expert tokens \(L\), are produced once after WAM inference and then stored as a \emph{KV cache}. During execution, the verifier only encodes the latest real observation \(O_t\) and performs lightweight attention against these cached tokens, which makes score computation efficient without rerunning the full WAM.

FFDC is implemented as an \(N\)-layer Transformer with a structured Boolean visibility matrix \(M\), where \(M(i,j)=1\) means token \(x_i\) can attend to token \(x_j\). The mask enforces causal interaction between future actions and future predicted dynamics. Specifically, besides attending to \(L\), \(O_t\), and \(\hat{O}_{t_p}\), each future visual token \(\hat{O}_{t_f}^{(j)}\) attends only to \(\{\hat{O}_{t_f}^{(\leq j)}, \hat{A}_{t}^{(\leq t+jr)}\}\), and each future action token attends only to \(\{\hat{O}_{t_f}^{(\leq j)}, \hat{A}_{t}^{(\leq t+jr)}\}\). To further reduce computation, this attention is applied with a local window over the temporally ordered future tokens, so each token interacts only with nearby aligned action/visual tokens rather than the full future sequence. This preserves temporal causality, avoids information leakage, and keeps the verifier lightweight.

Finally, a \texttt{[CLS]} token attends to the full visible sequence and aggregates the execution state into a compact representation. Its output is passed through an MLP head \(g_{\psi}\) to produce
\begin{equation}
z_t = g_{\psi}\!\left(\mathrm{FFDC}(X_t)_{\texttt{[CLS]}}\right),
\end{equation}
followed by
\begin{equation}
e_t = \sigma(z_t) \in [0,1],
\end{equation}
where a larger \(e_t\) indicates higher confidence that the future action segment \(\hat{A}_t\) remains valid under the latest real observation.

\subsection{Training strategy and dataset construction}
\label{sec:training strategy}
To improve trajectory coverage for long-horizon inference, we train WAM with a mixture-of-horizon sampling strategy. For an episode of length \(T\), we uniformly sample a conditioning timestep \(s \sim \mathcal{U}\{1,\dots,T\}\). Given horizon \(H\), the action and video indices are defined as \(\tau_i=\min(s+i,T)\), and \(\upsilon_j=\min(s+jr,T)\), where \(i=0,\dots,H-1\) and \(j=0,\dots,H/r-1\), which yield action and video sequences
\begin{equation}
A_s=[a_{\tau_0},\dots,a_{\tau_{H-1}}], \qquad
O_s=[o_{\upsilon_0},\dots,o_{\upsilon_{H/r-1}}].
\end{equation}
Out-of-range positions are padded by repeating the final valid action or frame. This allows late-stage states to serve as conditioning starts during training and reduces the bias toward early-episode prefixes.

For FFDC verifier training, we construct a binary dataset \(\mathcal{D}_{\mathrm{ver}}=\{(X,y)\}\), where \(X\) denotes the verifier input and \(y \in \{0,1\}\) indicates whether the future action segment is executable, with \(y=1\) for valid segments and \(y=0\) for failure-inducing ones. Positive samples are collected from demonstration data and a small number of successful rollouts. Negative samples are obtained from a small number of failed rollouts and from corrupted action segments synthesized from valid demonstrations. The data augmentation methods include temporal swap, gripper flip, late-stage Gaussian noise, and tail scaling. The temporal swap operator randomly swaps two pairs of actions within a horizon; gripper flip negates the designated gripper dimensions; late-stage Gaussian noise perturbs the second half of the sequence; and tail scaling shrinks a randomly sampled suffix by a random scale factor. Using the resulting dataset, the verifier is trained as a binary classifier with the loss

\begin{equation}
\mathcal{L}_{\mathrm{ver}}
=
-\Bigl[
y \log \sigma(z)
+
(1-y)\log \bigl(1-\sigma(z)\bigr)
\Bigr].
\end{equation}

\section{Experiments}
\label{sec:experiments}

\paragraph{Experimental setups.}
We evaluate FFDC-WAM on RoboTwin~\citep{chen2025robotwin} and on six real-world tasks. We implement all models in PyTorch with Motus~\citep{bi2025motus} as the WAM backbone. Backbone training uses four NVIDIA A100 GPUs (80GB each), verifier training uses one A100, and evaluation runs on a single A100.

We evaluate on 50 RoboTwin tasks~\citep{chen2025robotwin} under clean and randomized settings with 100 rollouts per task and setting, and on six real-world tasks using a Unitree G1D dual-arm robot with Dex-1 grippers. We report success rate (SR), average completion time ($T$), and WAM planning/replanning calls (Calls), excluding SGAC's additional verification inferences. We use a fixed split of five hard RoboTwin tasks and 45 easy tasks based on the Base-Motus SR, with overall SR averaged across all 50 tasks and per-task results in Appendix~\ref{appendix:more_results}.
\textbf{Base-Motus} uses a 16-action training and execution horizon, while \textbf{LC-16/32/48/64} execute different prefixes of a MoH-trained, 64-action backbone. \textbf{FFDC-WAM} and \textbf{SGAC}~\citep{so2026improving} share this backbone, adapting execution through future--reality verification and action-consistency checks, respectively.

\subsection{Main results}
\label{sec:main_results}

\begin{figure}[h]
  \centering
  \includegraphics[width=\linewidth, trim=25 190 20 25, clip]{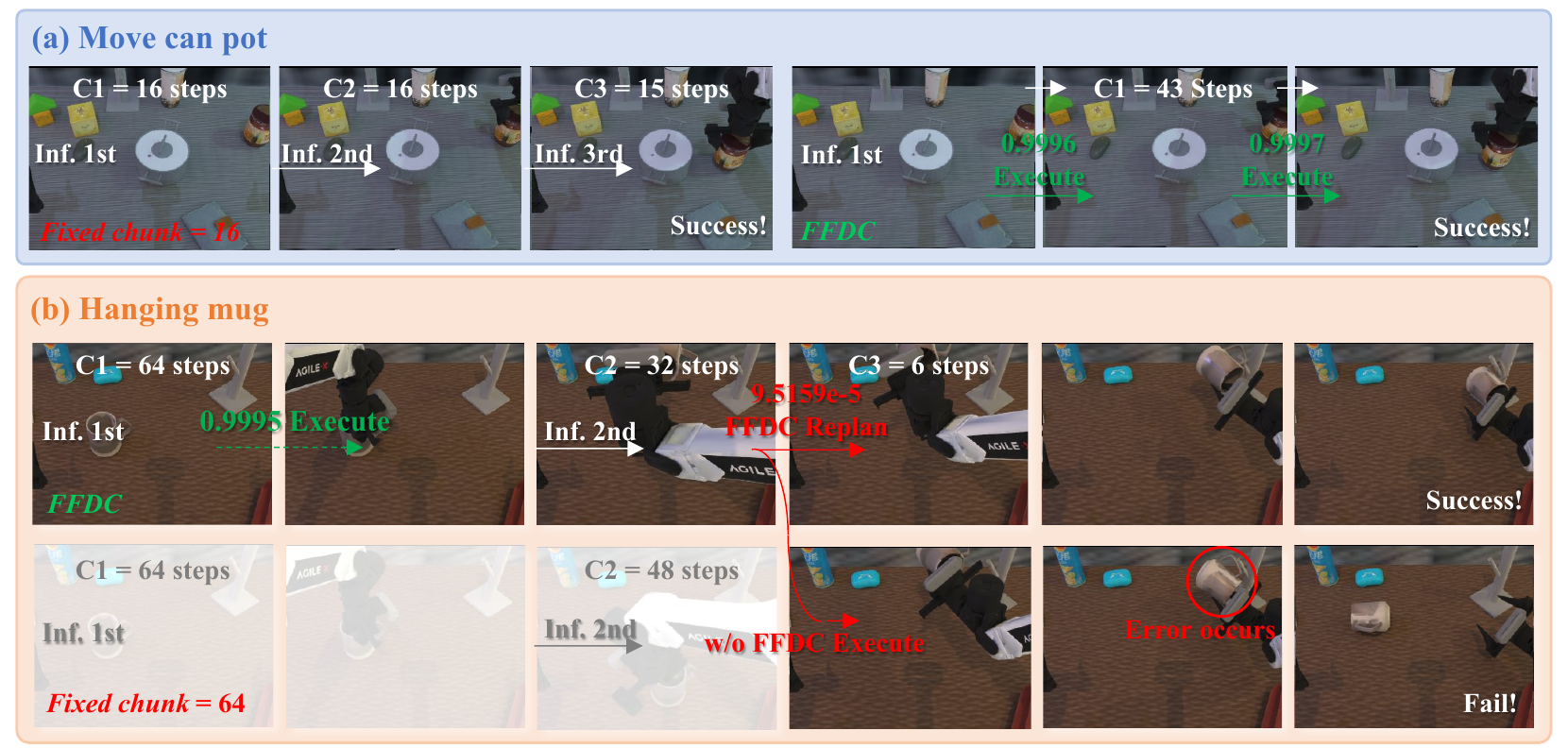}
  \caption{Adaptive execution on RoboTwin. (a) On \textit{Move Can Pot}, FFDC completes the illustrated rollout with one WAM inference instead of three for fixed 16-action execution. (b) On \textit{Hanging Mug}, it replans during the precision-critical stage and succeeds, whereas fixed 64-action execution fails. $C_i$ denotes an executed action chunk.}
  \label{fig:figure3_demo}
\end{figure}

\begin{table}[h]
\caption{RoboTwin results under (a) clean and (b) randomized conditions. Each panel reports the 45 easy tasks, five hard tasks, and task-weighted average over all 50 tasks. $^{\dagger}$Calls excludes SGAC's additional WAM inferences for verification.}
\label{tab:main_results}
\begin{center}
\footnotesize
\setlength{\tabcolsep}{3pt}
\renewcommand{\arraystretch}{1.06}
\textbf{(a) Clean}\par\smallskip
\begin{tabular*}{0.9\linewidth}{@{\extracolsep{\fill}}l*{9}{r}@{}}
\toprule
& \multicolumn{3}{c}{Easy} & \multicolumn{3}{c}{Hard} & \multicolumn{3}{c}{All tasks} \\
\cmidrule(lr){2-4}\cmidrule(lr){5-7}\cmidrule(l){8-10}
Method & SR (\%) & $T$ (s) & Calls & SR (\%) & $T$ (s) & Calls & SR (\%) & $T$ (s) & Calls \\
\midrule
Base-Motus & \textbf{90.98} & 20.4 & 5.22 & 57.80 & 29.5 & 7.71 & 87.66 & 21.3 & 5.47 \\
SGAC$^{\dagger}$ & 87.02 & 15.1 & 1.73 & 74.00 & 21.6 & 2.83 & 85.72 & 15.8 & 1.84 \\
Latent cosine gate & 87.24 & 13.0 & 1.68 & 72.40 & 19.3 & 2.69 & 85.76 & 13.6 & 1.78 \\
Pixel cosine gate & 87.96 & 30.5 & 1.95 & 75.60 & 35.2 & 2.53 & 86.72 & 31.0 & 2.01 \\
LC-16 & 86.82 & 18.3 & 4.57 & 64.40 & 26.7 & 6.84 & 84.58 & 19.1 & 4.80 \\
LC-32 & 89.38 & 13.4 & 2.57 & 70.60 & 18.3 & 3.56 & 87.50 & 13.8 & 2.67 \\
LC-48 & 89.73 & 11.4 & 1.70 & 67.40 & 16.1 & 2.52 & 87.50 & 11.8 & 1.79 \\
LC-64 & 88.98 & 10.8 & 1.62 & 73.80 & 14.0 & 2.03 & 87.46 & 11.1 & 1.66 \\
\midrule
FFDC-WAM & 90.33 & 12.9 & 1.62 & \textbf{76.00} & 18.8 & 2.60 & \textbf{88.90} & 13.5 & 1.72 \\
\bottomrule\end{tabular*}
\par\medskip
\textbf{(b) Randomized}\par\smallskip
\begin{tabular*}{0.9\linewidth}{@{\extracolsep{\fill}}l*{9}{r}@{}}
\toprule
& \multicolumn{3}{c}{Easy} & \multicolumn{3}{c}{Hard} & \multicolumn{3}{c}{All tasks} \\
\cmidrule(lr){2-4}\cmidrule(lr){5-7}\cmidrule(l){8-10}
Method & SR (\%) & $T$ (s) & Calls & SR (\%) & $T$ (s) & Calls & SR (\%) & $T$ (s) & Calls \\
\midrule
Base-Motus & 89.16 & 23.5 & 5.22 & 54.20 & 33.0 & 7.73 & 85.66 & 24.5 & 5.47 \\
SGAC$^{\dagger}$ & 85.69 & 18.9 & 1.89 & 72.80 & 23.4 & 2.47 & 84.40 & 19.4 & 1.95 \\
Latent cosine gate & 86.91 & 16.1 & 1.70 & 72.20 & 21.2 & 2.37 & 85.44 & 16.6 & 1.77 \\
Pixel cosine gate & 86.42 & 30.3 & 1.94 & 73.20 & 36.8 & 2.81 & 85.10 & 31.0 & 2.03 \\
LC-16 & 85.58 & 21.2 & 4.55 & 67.40 & 29.4 & 6.68 & 83.76 & 22.0 & 4.76 \\
LC-32 & 88.64 & 16.0 & 2.57 & 71.60 & 21.2 & 3.55 & 86.94 & 16.5 & 2.67 \\
LC-48 & 88.49 & 13.9 & 1.72 & 65.00 & 19.0 & 2.59 & 86.14 & 14.4 & 1.81 \\
LC-64 & 88.89 & 13.3 & 1.64 & 74.00 & 16.5 & 2.06 & 87.40 & 13.7 & 1.68 \\
\midrule
FFDC-WAM & \textbf{89.51} & 15.7 & 1.62 & \textbf{76.40} & 20.5 & 2.34 & \textbf{88.20} & 16.2 & 1.69 \\
\bottomrule\end{tabular*}
\end{center}
\end{table}

\paragraph{Results on RoboTwin.}
On RoboTwin (Table~\ref{tab:main_results}), FFDC-WAM achieves the highest average SR in both clean (88.90\%) and randomized (88.20\%) settings. Under randomization, it improves SR over Base-Motus by 2.54 percentage points while reducing completion time by 33.88\% and WAM calls by 69.10\%. Gains are most pronounced on hard tasks, where SR increases from 57.80\% to 76.00\% in clean conditions and from 54.20\% to 76.40\% under randomization; on easy tasks, it maintains comparable SR with shorter execution time. Compared with the same-backbone LC-64, FFDC-WAM improves hard-task SR by 2.20 and 2.40 percentage points, respectively.

Compared with adaptive baselines, FFDC-WAM exceeds SGAC's SR by 3.18 and 3.80 percentage points in clean and randomized settings, respectively, while completing tasks faster. Each FFDC check takes 0.015\,s on an A100, compared with 1.9\,s for SGAC, which requires fresh WAM inference. Latent cosine gating achieves comparable execution time but lower SR by 3.14 and 2.76 percentage points, respectively. Pixel-level gating also yields lower SR and incurs additional decoding cost, taking 1.16\,s per check and an average completion time of 31.0\,s in both settings. These results support jointly reasoning over predicted dynamics, real observations, actions, and task semantics for accurate and efficient execution verification.

\paragraph{Execution behavior.}
Figure~\ref{fig:figure3_demo} illustrates how verification changes the executed horizon. On \textit{Move Can Pot}, high confidence allows one 43-action chunk to complete the rollout, while fixed 16-action execution uses three WAM calls. On \textit{Hanging Mug}, FFDC executes 64 actions in the first chunk, then interrupts the second after 32 actions when confidence falls below the threshold; a final six-action chunk completes the task. As shown in the lower row, ignoring this confidence drop and continuing with fixed 64-action execution leads to failure. These examples illustrate both continued execution when a plan remains reliable and timely replanning when confidence deteriorates.

\subsection{Real-world experiments}
\label{sec:real_worlf_exp}

Table~\ref{tab:real-world} reports results on six manipulation tasks using a Unitree G1D dual-arm robot. FFDC-WAM achieves 75.00\% average SR, improving over Base-Motus-LC16 by 9.17 percentage points while reducing average completion time from 84.42 to 41.10\,s (51.31\%) and WAM calls from 25.44 to 6.60 (74.06\%). Success improves on all six tasks by 5--15 percentage points, with lower completion time and fewer calls on every task. Compared with fixed LC-64, FFDC-WAM improves average SR by 13.33 percentage points with only 2.33\,s and 0.22 additional calls. The largest gain is on Clean Table, from 55\% to 75\%, followed by 15-point gains on Put Cup and Stack Bowls. These results support combining efficient long-horizon execution with selective correction to improve real-world reliability.

\begin{figure}[h]
  \centering
  \includegraphics[width=\linewidth, trim=25 676 25 27, clip]{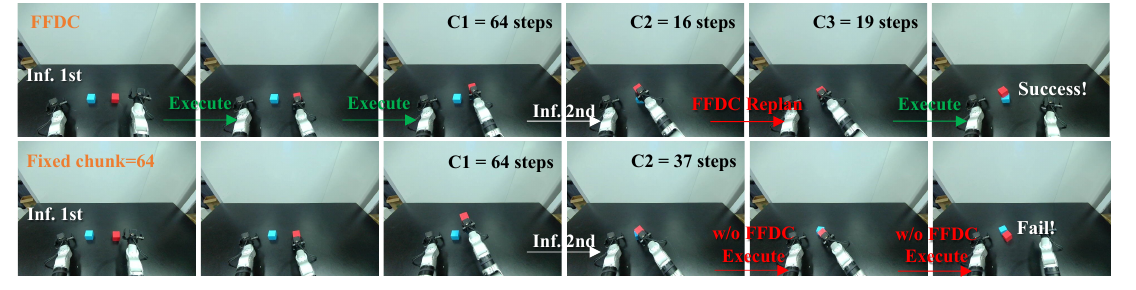}
  \caption{Real-world block manipulation on the G1D robot. FFDC executes a 64-action first chunk, interrupts the next chunk after 16 actions, and completes the illustrated rollout with a 19-action final chunk. The fixed 64-action baseline continues without this intermediate verification and fails.}
  \label{fig:fig4}
\end{figure}

Figure~\ref{fig:fig4} illustrates this behavior on a block-manipulation task. FFDC executes a full 64-action chunk, then interrupts the second chunk after 16 actions when confidence falls. Replanning produces a final 19-action chunk that completes the task, whereas the fixed LC-64 baseline continues its existing plan and fails. This example demonstrates adaptation within a single rollout: the robot retains long execution while the plan remains trusted and replans when the remaining actions become unreliable, consistent with the success gains and modest additional planning cost observed relative to LC-64.

\begin{table}[h]
\caption{Real-world performance on six manipulation tasks.}
\label{tab:real-world}
\begin{center}
\footnotesize
\setlength{\tabcolsep}{3pt}
\begin{tabular*}{0.9\linewidth}{@{\extracolsep{\fill}}l*{9}{c}@{}}
\toprule
& \multicolumn{3}{c}{Base-Motus-LC16}
& \multicolumn{3}{c}{LC-64}
& \multicolumn{3}{c}{FFDC-WAM} \\
\cmidrule(lr){2-4}\cmidrule(lr){5-7}\cmidrule(lr){8-10}
Task & SR (\%) & $T$ (s) & Calls
     & SR (\%) & $T$ (s) & Calls
     & SR (\%) & $T$ (s) & Calls \\
\midrule
Bowl Pouring & 65 & 70.94  & 19.10 & 65 & 35.67 & 5.15 & \textbf{75} & 39.19 & 5.65 \\
Clean Table  & 60 & 113.27 & 36.30 & 55 & 46.36 & 8.75 & \textbf{75} & 47.52 & 9.00 \\
Put Block    & 70 & 66.95  & 17.80 & 70 & 34.10 & 4.90 & \textbf{80} & 36.55 & 4.95 \\
Put Cup      & 70 & 69.24  & 20.00 & 65 & 35.04 & 4.95 & \textbf{80} & 37.40 & 5.05 \\
Stack Blocks & 55 & 68.38  & 19.70 & 50 & 34.59 & 4.80 & \textbf{60} & 37.06 & 4.90 \\
Stack Bowls  & 75 & 117.72 & 39.75 & 65 & 46.87 & 9.70 & \textbf{80} & 48.90 & 10.05 \\
\midrule
Avg. & 65.83 & 84.42 & 25.44
     & 61.67 & 38.77 & 6.38
     & \textbf{75.00} & 41.10 & 6.60 \\
\bottomrule
\end{tabular*}
\end{center}
\end{table}

\subsection{Ablation study}
\label{sec:ablation}

We conduct ablations on the five hard RoboTwin tasks under the more challenging randomized setting (Table~\ref{tab:ablation_results}). In the variant names, \textbf{MoH} denotes Mixture-of-Horizon Training. The two \textbf{w/o MoH} variants use a backbone trained without MoH, with either fixed 64-action execution (\textbf{LC-64}) or adaptive verification (\textbf{FFDC}). The input ablations remove instruction-conditioned understanding tokens (\textbf{Und}), predicted visual tokens (\textbf{Pred}), the current real observation (\textbf{Real}), or predicted actions (\textbf{Action}). The decision threshold is denoted by $\tau$, with $\tau=0.5$ as the default. The $\tau=0.25$ and $\tau=0.75$ variants change only this threshold to evaluate more permissive and stricter continuation criteria, respectively. \textbf{FFDC-WAM} denotes the complete model with the default threshold. BlkRank, HangMug, PlacePad, PutCab, and ScanObj denote Blocks Ranking Size, Hanging Mug, Place Mouse Pad, Put Object in Cabinet, and Scan Object, respectively.

\begin{table}[h]
\centering
\caption{Ablations on five hard RoboTwin tasks under randomization: (a) verifier inputs; (b) MoH training and decision threshold. SR is in percent; $T$ is in seconds. Average denotes the mean across the five tasks. Bold marks the highest SR in each row within each panel.}
\label{tab:ablation_results}
\vspace{0.15in}

\footnotesize
\setlength{\tabcolsep}{2pt}
\renewcommand{\arraystretch}{1.12}

\textbf{(a) Verifier inputs}\par\smallskip
\begin{tabular*}{\linewidth}{@{\extracolsep{\fill}}l*{15}{r}@{}}
\toprule
& \multicolumn{3}{c}{w/o Und}
& \multicolumn{3}{c}{w/o Pred}
& \multicolumn{3}{c}{w/o Real}
& \multicolumn{3}{c}{w/o Action}
& \multicolumn{3}{c}{FFDC-WAM} \\
\cmidrule(lr){2-4}
\cmidrule(lr){5-7}
\cmidrule(lr){8-10}
\cmidrule(lr){11-13}
\cmidrule(l){14-16}
Task
& SR & $T$ (s) & Calls
& SR & $T$ (s) & Calls
& SR & $T$ (s) & Calls
& SR & $T$ (s) & Calls
& SR & $T$ (s) & Calls \\
\midrule
BlkRank
& 91.0 & 26.1 & 3.34
& 85.0 & 24.9 & 3.11
& 86.0 & 27.8 & 3.70
& 90.0 & 25.5 & 3.31
& \textbf{93.0} & 25.3 & 3.27 \\
HangMug
& 37.0 & 29.2 & 4.16
& 32.0 & 27.8 & 4.03
& 38.0 & 27.3 & 3.92
& 33.0 & 28.9 & 4.24
& \textbf{44.0} & 27.5 & 3.96 \\
PlacePad
& 86.0 & 12.5 & 1.23
& 84.0 & 12.4 & 1.24
& 87.0 & 12.3 & 1.20
& \textbf{88.0} & 13.1 & 1.43
& \textbf{88.0} & 12.4 & 1.27 \\
PutCab
& 72.0 & 26.6 & 3.49
& 71.0 & 21.9 & 2.01
& 68.0 & 21.7 & 2.00
& 69.0 & 22.1 & 2.00
& \textbf{73.0} & 21.7 & 2.00 \\
ScanObj
& \textbf{88.0} & 16.0 & 1.25
& 86.0 & 16.8 & 1.36
& 83.0 & 15.9 & 1.23
& 87.0 & 15.6 & 1.22
& 84.0 & 15.7 & 1.20 \\
\midrule
Average
& 74.8 & 22.1 & 2.69
& 71.6 & 20.8 & 2.35
& 72.4 & 21.0 & 2.41
& 73.4 & 21.0 & 2.44
& \textbf{76.4} & 20.5 & 2.34 \\
\bottomrule
\end{tabular*}

\par\medskip

\textbf{(b) MoH training and decision threshold}\par\smallskip
\begin{tabular*}{\linewidth}{@{\extracolsep{\fill}}l*{15}{r}@{}}
\toprule
& \multicolumn{3}{c}{\shortstack{w/o MoH\\LC-64}}
& \multicolumn{3}{c}{\shortstack{w/o MoH\\FFDC}}
& \multicolumn{3}{c}{$\tau=0.25$}
& \multicolumn{3}{c}{$\tau=0.75$}
& \multicolumn{3}{c}{FFDC-WAM} \\
\cmidrule(lr){2-4}
\cmidrule(lr){5-7}
\cmidrule(lr){8-10}
\cmidrule(lr){11-13}
\cmidrule(l){14-16}
Task
& SR & $T$ (s) & Calls
& SR & $T$ (s) & Calls
& SR & $T$ (s) & Calls
& SR & $T$ (s) & Calls
& SR & $T$ (s) & Calls \\
\midrule
BlkRank
& 77.0 & 23.8 & 3.49
& 90.0 & 26.7 & 3.31
& 91.0 & 25.6 & 3.18
& 91.0 & 25.7 & 3.23
& \textbf{93.0} & 25.3 & 3.27 \\
HangMug
& 40.0 & 19.8 & 2.00
& \textbf{44.0} & 25.9 & 3.45
& 39.0 & 27.8 & 3.90
& 42.0 & 27.7 & 4.00
& \textbf{44.0} & 27.5 & 3.96 \\
PlacePad
& 87.0 & 9.9 & 1.06
& 67.0 & 15.1 & 1.59
& \textbf{89.0} & 12.6 & 1.26
& 88.0 & 13.4 & 1.44
& 88.0 & 12.4 & 1.27 \\
PutCab
& 48.0 & 21.3 & 2.04
& \textbf{75.0} & 22.8 & 2.03
& 69.0 & 22.6 & 2.01
& 68.0 & 22.2 & 2.04
& 73.0 & 21.7 & 2.00 \\
ScanObj
& \textbf{88.0} & 13.1 & 1.13
& 81.0 & 15.3 & 1.19
& 85.0 & 16.9 & 1.28
& 87.0 & 15.6 & 1.14
& 84.0 & 15.7 & 1.20 \\
\midrule
Average
& 68.0 & 17.6 & 1.94
& 71.4 & 21.2 & 2.31
& 74.6 & 21.1 & 2.33
& 75.2 & 20.9 & 2.37
& \textbf{76.4} & 20.5 & 2.34 \\
\bottomrule
\end{tabular*}
\end{table}

\paragraph{Verifier analysis.}
The complete verifier achieves 76.4\% average SR. Removing Pred, Real, Action, or Und reduces SR to 71.6\%, 72.4\%, 73.4\%, and 74.8\%, respectively, while increasing completion time. Predicted visuals contribute the largest average gain, with their removal also reducing Hanging Mug SR from 44\% to 32\%. These results support jointly considering all four inputs. Offline evaluation on approximately 2,000 task-and-seed-matched FFDC and LC-64 episodes, labeled by final episode outcomes, yields an AUROC of 0.786 and an AUPRC of 0.761. Matched LC-64 rollouts without verification fail in 77.8\% of cases associated with a replan prediction, versus 31.6\% with a continue prediction, supporting the verifier's ability to identify unreliable continuation.

\paragraph{MoH training and decision threshold.}
MoH improves SR from 68.0\% to 74.0\% for fixed LC-64 and from 71.4\% to 76.4\% for FFDC-WAM; the MoH-trained LC-64 results are reported in Table~\ref{tab:main_results}. Verification improves SR both with and without MoH, indicating complementary contributions from long-horizon training and online correction. Setting $\tau$ to 0.25 or 0.75 yields 74.6\% or 75.2\% SR, versus 76.4\% at the default. Across these thresholds, completion time remains within 20.5--21.1\,s and calls within 2.33--2.37, showing limited sensitivity over the tested range.

\section{Conclusion}
Inspired by how humans compare predicted future feedback with actual observation during physical interaction, we reformulate adaptive WAM execution as a \emph{future--reality verification} problem. Instead of using a fixed execution horizon, our method asks a more fundamental question: \emph{whether the WAM's imagined future can still be trusted during rollout}. To this end, we propose \textbf{FFDC-WAM}, where the lightweight \emph{Future Forward Dynamics Causal Attention} (FFDC) verifier jointly models temporally aligned interactions among predicted future actions, predicted visual dynamics, real observations, and language instructions to detect unreliable future execution.

This design enables adaptive trust in WAM imagination: the robot can continue execution when the predicted future remains reliable, and trigger replanning once future--reality consistency breaks down. In this way, FFDC-WAM moves WAM execution beyond fixed-horizon chunking toward a more principled reliability-aware control strategy, improving both efficiency and robustness across simulation and real-world settings. More broadly, our results suggest that the key to effective WAM deployment is not selecting a single execution length, but endowing the system with the ability to verify its own imagined future online. A current limitation is that FFDC is trained with binary supervision derived from successful, failed, and synthetically corrupted segments, which may not cover the full diversity of real-world execution deviations. Extending the verifier to learn from richer failure modes and more diverse real-world data is an important direction for future work.

\subsection*{AI use statement}

We used generative AI tools to assist with language polishing, including improving grammar, clarity, and readability. All AI-assisted edits were reviewed and revised by the authors. We take full responsibility for the final content of this paper, including its claims, results, and conclusions.

\bibliography{iclr2027_conference}
\bibliographystyle{iclr2027_conference}

\appendix
\section{Appendix}

\subsection{More results}
\label{appendix:more_results}

\begin{table}[h]
\caption{Results on hard tasks under random conditions.}
\label{tab:hard_random}
\begin{center}
\footnotesize
\setlength{\tabcolsep}{3pt}
\renewcommand{\arraystretch}{1.15}
\resizebox{0.9\linewidth}{!}{%
\begin{tabular}{@{}l*{12}{c}@{}}
\toprule
& \multicolumn{2}{c}{Base-Motus}
& \multicolumn{2}{c}{LC-16}
& \multicolumn{2}{c}{LC-32}
& \multicolumn{2}{c}{LC-48}
& \multicolumn{2}{c}{LC-64}
& \multicolumn{2}{c}{FFDC-WAM} \\
\cmidrule(lr){2-3}
\cmidrule(lr){4-5}
\cmidrule(lr){6-7}
\cmidrule(lr){8-9}
\cmidrule(lr){10-11}
\cmidrule(l){12-13}
Task
& SR (\%) & T (s)
& SR (\%) & T (s)
& SR (\%) & T (s)
& SR (\%) & T (s)
& SR (\%) & T (s)
& SR (\%) & T (s) \\
\midrule
BlkRank
& 62 & 49.1 & 66 & 43.1 & 78 & 30.0
& 62 & 26.7 & 86 & 21.5 & \textbf{93} & 25.3 \\
HangMug
& 38 & 36.2 & 42 & 31.8 & 35 & 24.6
& 26 & 22.0 & 33 & 19.0 & \textbf{44} & 27.5 \\
PlacePad
& 64 & 23.5 & 76 & 19.9 & 85 & 13.7
& 86 & 11.5 & \textbf{90} & 10.5 & 88 & 12.4 \\
PutCab
& 53 & 32.5 & 67 & 29.7 & 67 & 21.0
& 66 & 18.7 & \textbf{75} & 18.0 & 73 & 21.7 \\
ScanObj
& 54 & 23.9 & 86 & 22.6 & \textbf{93} & 16.7
& 85 & 16.0 & 86 & 13.8 & 84 & 15.7 \\
\midrule
Average
& 54.2 & 33.0 & 67.4 & 29.4 & 71.6 & 21.2
& 65.0 & 19.0 & 74.0 & 16.5 & \textbf{76.4} & 20.5 \\
\bottomrule
\end{tabular}%
}    
\end{center}
\end{table}

\begin{table}[h]
\caption{Results on hard tasks under clean conditions.}
\label{tab:hard_clean}
\begin{center}
\footnotesize
\setlength{\tabcolsep}{3pt}
\renewcommand{\arraystretch}{1.15}
\resizebox{0.9\linewidth}{!}{%
\begin{tabular}{@{}l*{12}{c}@{}}
\toprule
& \multicolumn{2}{c}{Base-Motus}
& \multicolumn{2}{c}{LC-16}
& \multicolumn{2}{c}{LC-32}
& \multicolumn{2}{c}{LC-48}
& \multicolumn{2}{c}{LC-64}
& \multicolumn{2}{c}{FFDC-WAM} \\
\cmidrule(lr){2-3}
\cmidrule(lr){4-5}
\cmidrule(lr){6-7}
\cmidrule(lr){8-9}
\cmidrule(lr){10-11}
\cmidrule(l){12-13}
Task
& SR (\%) & T (s)
& SR (\%) & T (s)
& SR (\%) & T (s)
& SR (\%) & T (s)
& SR (\%) & T (s)
& SR (\%) & T (s) \\
\midrule
BlkRank
& 78 & 40.3 & 65 & 40.9 & 78 & 26.2
& 63 & 21.5 & 81 & 18.5 & \textbf{84} & 25.0 \\
HangMug
& 27 & 30.8 & 27 & 29.5 & 23 & 22.2
& 21 & 19.7 & 26 & 16.8 & \textbf{30} & 26.5 \\
PlacePad
& 58 & 28.2 & 80 & 18.9 & 85 & 11.5
& 89 & 8.0 & \textbf{95} & 7.5 & 94 & 10.1 \\
PutCab
& 62 & 29.3 & 60 & 26.1 & 76 & 18.2
& 73 & 16.2 & 79 & 16.1 & \textbf{80} & 19.2 \\
ScanObj
& 64 & 18.7 & 90 & 18.1 & 91 & 13.3
& 91 & 15.1 & 88 & 11.3 & \textbf{92} & 13.2 \\
\midrule
Average
& 57.8 & 29.5 & 64.4 & 26.7 & 70.6 & 18.3
& 67.4 & 16.1 & 73.8 & 14.0 & \textbf{76.0} & 18.8 \\
\bottomrule
\end{tabular}%
}
\end{center}
\end{table}

\begin{table}[h]
\caption{Results on easy tasks on Clean Settings.}
\label{tab:easy_clean}
\begin{center}
\scriptsize
\setlength{\tabcolsep}{2pt}
\renewcommand{\arraystretch}{1.08}
\resizebox{\linewidth}{!}{%
\begin{tabular}{@{}l*{12}{c}@{}}
\toprule
& \multicolumn{2}{c}{Base-Motus}
& \multicolumn{2}{c}{LC-16}
& \multicolumn{2}{c}{LC-32}
& \multicolumn{2}{c}{LC-48}
& \multicolumn{2}{c}{LC-64}
& \multicolumn{2}{c}{FFDC-WAM} \\
\cmidrule(lr){2-3}
\cmidrule(lr){4-5}
\cmidrule(lr){6-7}
\cmidrule(lr){8-9}
\cmidrule(lr){10-11}
\cmidrule(l){12-13}
Task
& SR (\%) & T (s)
& SR (\%) & T (s)
& SR (\%) & T (s)
& SR (\%) & T (s)
& SR (\%) & T (s)
& SR (\%) & T (s) \\
\midrule
adjust\_bottle
& 91 & 13.1 & \textbf{100} & 12.2 & \textbf{100} & 9.8 & \textbf{100} & 7.4 & 88 & 7.5 & \textbf{100} & 8.7 \\
beat\_block\_hammer
& \textbf{97} & 13.3 & 94 & 13.6 & 84 & 10.5 & 85 & 8.5 & 83 & 8.8 & 90 & 11.6 \\
blocks\_ranking\_rgb
& \textbf{100} & 36.3 & 86 & 40.3 & 98 & 24.3 & 81 & 22.0 & 94 & 18.2 & 98 & 21.2 \\
click\_alarmclock
& \textbf{100} & 7.7 & \textbf{100} & 7.6 & \textbf{100} & 5.3 & \textbf{100} & 5.3 & 99 & 5.4 & \textbf{100} & 6.1 \\
click\_bell
& \textbf{100} & 7.8 & \textbf{100} & 7.6 & \textbf{100} & 5.2 & \textbf{100} & 5.2 & 97 & 5.2 & \textbf{100} & 5.9 \\
dump\_bin\_bigbin
& \textbf{98} & 18.1 & 97 & 16.8 & 94 & 12.8 & 96 & 10.6 & 92 & 10.7 & \textbf{98} & 12.0 \\
grab\_roller
& \textbf{100} & 9.2 & \textbf{100} & 9.1 & \textbf{100} & 6.4 & \textbf{100} & 6.5 & 94 & 6.6 & \textbf{100} & 7.4 \\
handover\_block
& \textbf{87} & 25.3 & 44 & 29.8 & 73 & 17.6 & 68 & 13.7 & 68 & 13.9 & 60 & 20.1 \\
handover\_mic
& 84 & 21.7 & 95 & 19.8 & 99 & 15.6 & 96 & 13.5 & 99 & 13.5 & \textbf{100} & 15.4 \\
lift\_pot
& \textbf{99} & 16.8 & 92 & 11.2 & 91 & 11.4 & 93 & 8.5 & 97 & 8.3 & 93 & 10.4 \\
move\_can\_pot
& 37 & 25.4 & \textbf{86} & 13.5 & 82 & 10.6 & 85 & 8.5 & 85 & 8.4 & 82 & 9.9 \\
move\_pillbottle\_pad
& 96 & 16.0 & 91 & 12.2 & \textbf{98} & 9.5 & 96 & 7.6 & 97 & 7.2 & 96 & 8.5 \\
move\_playingcard\_away
& 99 & 12.9 & 99 & 11.2 & \textbf{100} & 8.6 & 99 & 6.4 & 97 & 6.4 & 99 & 7.5 \\
move\_stapler\_pad
& 82 & 32.8 & 85 & 22.8 & 84 & 15.1 & \textbf{90} & 14.6 & 82 & 12.0 & 88 & 15.4 \\
open\_laptop
& 94 & 16.5 & 98 & 13.9 & 97 & 11.0 & 98 & 10.2 & \textbf{100} & 9.3 & 98 & 11.1 \\
open\_microwave
& \textbf{92} & 46.1 & 75 & 26.7 & 54 & 16.5 & 71 & 14.9 & 79 & 12.9 & 71 & 17.0 \\
pick\_diverse\_bottles
& \textbf{89} & 12.4 & 71 & 12.4 & 67 & 10.2 & 68 & 8.0 & 69 & 8.0 & 70 & 9.4 \\
pick\_dual\_bottles
& \textbf{95} & 12.2 & 75 & 12.3 & 74 & 10.0 & 73 & 7.7 & 73 & 7.7 & 74 & 9.3 \\
place\_a2b\_left
& 88 & 22.7 & 93 & 13.3 & \textbf{97} & 10.0 & 96 & 8.1 & 94 & 7.9 & 96 & 9.1 \\
place\_a2b\_right
& 84 & 24.1 & 94 & 12.7 & 97 & 10.0 & \textbf{99} & 8.2 & 92 & 8.0 & 98 & 9.3 \\
place\_bread\_basket
& 90 & 22.1 & 92 & 20.5 & 95 & 15.5 & 85 & 14.1 & 91 & 13.0 & \textbf{96} & 14.9 \\
place\_bread\_skillet
& 86 & 15.7 & 86 & 14.3 & 89 & 12.1 & \textbf{92} & 9.7 & 91 & 10.1 & 90 & 11.9 \\
place\_burger\_fries
& 98 & 22.5 & 94 & 24.4 & 93 & 17.6 & 96 & 15.5 & \textbf{99} & 15.2 & 96 & 17.3 \\
place\_can\_basket
& 82 & 34.5 & 57 & 30.0 & 75 & 21.1 & \textbf{87} & 18.9 & 78 & 18.9 & 80 & 22.4 \\
place\_cans\_plasticbox
& 98 & 23.9 & 69 & 27.5 & 99 & 18.4 & \textbf{100} & 16.1 & 98 & 15.8 & \textbf{100} & 18.0 \\
place\_container\_plate
& 98 & 12.7 & 97 & 12.2 & \textbf{100} & 9.8 & 99 & 7.5 & 98 & 7.5 & 98 & 8.7 \\
place\_dual\_shoes
& \textbf{93} & 25.3 & 76 & 26.2 & 81 & 19.2 & 77 & 15.5 & 78 & 15.5 & 77 & 18.6 \\
place\_empty\_cup
& 99 & 15.8 & \textbf{100} & 12.8 & 99 & 10.5 & 96 & 9.9 & \textbf{100} & 8.1 & 99 & 9.9 \\
place\_fan
& 94 & 17.2 & 87 & 14.7 & 94 & 11.8 & 92 & 9.6 & 93 & 8.7 & \textbf{96} & 10.6 \\
place\_object\_basket
& \textbf{84} & 31.2 & 56 & 27.9 & 77 & 21.2 & 60 & 18.7 & 70 & 17.3 & 71 & 19.7 \\
place\_object\_scale
& 81 & 16.3 & 78 & 14.0 & 86 & 11.0 & \textbf{94} & 8.3 & 91 & 8.3 & 92 & 9.5 \\
place\_object\_stand
& 98 & 12.1 & 92 & 12.3 & 96 & 9.9 & 92 & 7.5 & \textbf{99} & 7.3 & 95 & 8.6 \\
place\_phone\_stand
& 89 & 12.6 & 93 & 12.3 & 95 & 10.7 & 93 & 8.0 & \textbf{96} & 7.8 & \textbf{96} & 9.1 \\
place\_shoe
& \textbf{100} & 15.6 & 99 & 15.5 & 99 & 11.2 & 99 & 9.3 & 99 & 8.4 & 98 & 10.0 \\
press\_stapler
& \textbf{96} & 9.5 & \textbf{96} & 10.8 & 93 & 6.6 & 93 & 6.7 & 90 & 6.8 & 93 & 7.6 \\
put\_bottles\_dustbin
& \textbf{88} & 48.6 & 77 & 45.6 & 78 & 32.6 & 80 & 26.7 & 75 & 26.4 & 78 & 36.7 \\
rotate\_qrcode
& 83 & 17.6 & 88 & 14.3 & 88 & 11.0 & \textbf{92} & 9.0 & \textbf{92} & 8.3 & \textbf{92} & 9.9 \\
shake\_bottle\_horizontally
& \textbf{100} & 10.0 & 99 & 9.4 & \textbf{100} & 7.3 & \textbf{100} & 6.4 & 97 & 6.5 & 99 & 7.4 \\
shake\_bottle
& \textbf{100} & 9.8 & \textbf{100} & 9.4 & \textbf{100} & 7.2 & \textbf{100} & 6.4 & \textbf{100} & 6.5 & \textbf{100} & 7.5 \\
stack\_blocks\_three
& \textbf{86} & 37.1 & 82 & 34.5 & 77 & 23.8 & 85 & 21.7 & 80 & 18.4 & 84 & 21.4 \\
stack\_blocks\_two
& \textbf{100} & 24.4 & 99 & 22.4 & 96 & 15.9 & \textbf{100} & 14.1 & 97 & 12.6 & \textbf{100} & 14.8 \\
stack\_bowls\_three
& 71 & 41.5 & 70 & 43.3 & 75 & 30.3 & \textbf{82} & 26.2 & 63 & 24.8 & 77 & 27.2 \\
stack\_bowls\_two
& 96 & 26.1 & 95 & 24.2 & 98 & 17.4 & 98 & 15.5 & \textbf{100} & 14.8 & 94 & 17.3 \\
stamp\_seal
& \textbf{98} & 16.7 & 90 & 14.6 & 90 & 11.5 & 93 & 8.1 & 88 & 8.0 & 92 & 10.5 \\
turn\_switch
& \textbf{74} & 9.4 & 60 & 11.8 & 60 & 6.2 & 59 & 6.3 & 62 & 6.5 & 61 & 7.0 \\
\midrule
Average
& \textbf{90.98} & 20.4
& 86.82 & 18.3
& 89.38 & 13.4
& 89.73 & 11.4
& 88.98 & 10.8
& 90.33 & 12.9 \\
\bottomrule
\end{tabular}%
}
\end{center}
\end{table}

\begin{table}[h]
\caption{Results on easy tasks under random conditions.}
\label{tab:easy_random}
\begin{center}
\scriptsize
\setlength{\tabcolsep}{2pt}
\renewcommand{\arraystretch}{1.08}
\resizebox{\linewidth}{!}{%
\begin{tabular}{@{}l*{12}{c}@{}}
\toprule
& \multicolumn{2}{c}{Base-Motus}
& \multicolumn{2}{c}{LC-16}
& \multicolumn{2}{c}{LC-32}
& \multicolumn{2}{c}{LC-48}
& \multicolumn{2}{c}{LC-64}
& \multicolumn{2}{c}{FFDC-WAM} \\
\cmidrule(lr){2-3}
\cmidrule(lr){4-5}
\cmidrule(lr){6-7}
\cmidrule(lr){8-9}
\cmidrule(lr){10-11}
\cmidrule(l){12-13}
Task
& SR (\%) & T (s)
& SR (\%) & T (s)
& SR (\%) & T (s)
& SR (\%) & T (s)
& SR (\%) & T (s)
& SR (\%) & T (s) \\
\midrule
adjust\_bottle
& 87 & 15.5 & 98 & 14.7 & 98 & 12.2 & \textbf{99} & 9.6 & 93 & 9.8 & 98 & 11.2 \\
beat\_block\_hammer
& 90 & 15.4 & \textbf{96} & 15.2 & 92 & 12.1 & 95 & 10.5 & 91 & 10.4 & 94 & 11.7 \\
blocks\_ranking\_rgb
& 93 & 44.6 & 91 & 41.3 & 94 & 28.6 & 80 & 27.5 & 94 & 21.1 & \textbf{97} & 24.7 \\
click\_alarmclock
& 99 & 10.3 & \textbf{100} & 10.1 & \textbf{100} & 7.8 & \textbf{100} & 7.6 & 98 & 7.8 & \textbf{100} & 8.5 \\
click\_bell
& \textbf{100} & 10.2 & \textbf{100} & 10.3 & \textbf{100} & 7.5 & \textbf{100} & 7.5 & 97 & 7.6 & \textbf{100} & 8.3 \\
dump\_bin\_bigbin
& 94 & 22.2 & \textbf{95} & 20.4 & 92 & 14.3 & 88 & 12.3 & 92 & 12.7 & \textbf{95} & 14.3 \\
grab\_roller
& \textbf{100} & 12.3 & \textbf{100} & 11.5 & \textbf{100} & 8.8 & \textbf{100} & 8.8 & 99 & 9.0 & \textbf{100} & 10.0 \\
handover\_block
& \textbf{74} & 37.7 & 38 & 30.2 & 57 & 19.7 & 57 & 16.6 & 61 & 16.7 & 57 & 19.5 \\
handover\_mic
& 69 & 26.5 & 97 & 24.6 & \textbf{99} & 19.3 & 98 & 17.2 & 98 & 17.5 & 98 & 19.6 \\
lift\_pot
& 96 & 17.0 & 92 & 14.2 & 91 & 14.1 & 97 & 10.5 & \textbf{98} & 10.6 & \textbf{98} & 12.9 \\
move\_can\_pot
& 76 & 23.3 & 74 & 15.6 & \textbf{85} & 13.0 & 83 & 10.6 & 80 & 10.7 & 84 & 12.4 \\
move\_pillbottle\_pad
& 95 & 19.9 & 94 & 14.6 & 98 & 11.8 & \textbf{99} & 9.6 & 98 & 9.5 & 98 & 10.9 \\
move\_playingcard\_away
& 95 & 15.6 & 98 & 13.8 & \textbf{99} & 11.5 & \textbf{99} & 8.7 & 97 & 8.9 & \textbf{99} & 10.2 \\
move\_stapler\_pad
& 82 & 29.5 & 77 & 28.5 & 82 & 19.5 & 86 & 16.2 & \textbf{88} & 15.5 & 85 & 18.9 \\
open\_laptop
& 90 & 19.6 & 97 & 18.4 & 97 & 14.3 & 98 & 13.5 & \textbf{99} & 11.9 & 97 & 14.3 \\
open\_microwave
& \textbf{92} & 50.4 & 60 & 33.9 & 55 & 17.8 & 59 & 19.9 & 56 & 18.0 & 60 & 20.8 \\
pick\_diverse\_bottles
& \textbf{85} & 14.7 & 65 & 14.4 & 63 & 12.3 & 61 & 9.7 & 69 & 9.8 & 66 & 11.6 \\
pick\_dual\_bottles
& \textbf{88} & 14.8 & 63 & 14.4 & 59 & 12.2 & 58 & 9.5 & 64 & 9.6 & 63 & 11.2 \\
place\_a2b\_left
& 84 & 22.9 & 92 & 15.3 & 95 & 12.6 & 92 & 10.3 & \textbf{97} & 10.2 & 93 & 11.8 \\
place\_a2b\_right
& 80 & 25.3 & 94 & 15.9 & 97 & 12.6 & 97 & 10.7 & 94 & 10.3 & \textbf{98} & 11.9 \\
place\_bread\_basket
& 93 & 23.7 & 89 & 23.4 & 91 & 16.8 & 82 & 15.7 & 91 & 15.1 & \textbf{96} & 17.0 \\
place\_bread\_skillet
& 84 & 18.6 & 83 & 18.2 & 93 & 15.0 & 84 & 12.2 & \textbf{94} & 12.2 & 90 & 13.6 \\
place\_burger\_fries
& \textbf{99} & 25.6 & 91 & 27.2 & 91 & 20.2 & 96 & 18.9 & 90 & 18.0 & 90 & 20.9 \\
place\_can\_basket
& \textbf{76} & 37.7 & 66 & 32.5 & 71 & 28.3 & \textbf{76} & 23.6 & 67 & 25.3 & 72 & 26.5 \\
place\_cans\_plasticbox
& 95 & 27.3 & 79 & 33.1 & \textbf{100} & 21.0 & 99 & 18.7 & 94 & 18.5 & 97 & 20.7 \\
place\_container\_plate
& 98 & 15.1 & \textbf{100} & 14.9 & 98 & 12.0 & 99 & 9.8 & 98 & 9.8 & \textbf{100} & 11.3 \\
place\_dual\_shoes
& \textbf{90} & 29.4 & 65 & 27.6 & 83 & 22.2 & 74 & 18.7 & 76 & 18.5 & 77 & 24.1 \\
place\_empty\_cup
& \textbf{100} & 19.4 & 98 & 16.3 & 99 & 12.8 & 99 & 12.5 & 98 & 10.4 & 99 & 12.5 \\
place\_fan
& 80 & 18.1 & 88 & 17.8 & 91 & 13.6 & 96 & 11.5 & \textbf{99} & 11.1 & 94 & 12.6 \\
place\_object\_basket
& 81 & 38.7 & 73 & 33.0 & 78 & 26.0 & 73 & 23.0 & \textbf{85} & 23.5 & 77 & 26.1 \\
place\_object\_scale
& 87 & 20.0 & 73 & 16.3 & 89 & 14.0 & \textbf{96} & 11.1 & 95 & 10.4 & 94 & 11.7 \\
place\_object\_stand
& 97 & 15.2 & 92 & 14.8 & 98 & 12.1 & \textbf{99} & 9.6 & 95 & 9.8 & \textbf{99} & 11.2 \\
place\_phone\_stand
& 81 & 14.9 & 91 & 15.1 & 90 & 12.2 & 92 & 9.9 & \textbf{93} & 10.1 & 90 & 11.0 \\
place\_shoe
& 98 & 18.9 & 98 & 18.4 & 99 & 13.1 & \textbf{100} & 11.0 & 98 & 10.5 & \textbf{100} & 12.3 \\
press\_stapler
& 97 & 13.3 & \textbf{98} & 12.4 & \textbf{98} & 9.8 & \textbf{98} & 10.1 & 97 & 9.8 & \textbf{98} & 10.9 \\
put\_bottles\_dustbin
& \textbf{77} & 52.0 & 70 & 48.5 & 66 & 32.6 & 74 & 28.1 & 65 & 24.7 & 73 & 36.7 \\
rotate\_qrcode
& 75 & 22.8 & 86 & 17.6 & \textbf{92} & 13.6 & 89 & 11.2 & 90 & 10.6 & 90 & 12.4 \\
shake\_bottle\_horizontally
& 95 & 12.6 & 98 & 12.1 & 98 & 9.9 & 98 & 8.9 & \textbf{100} & 8.8 & 99 & 10.0 \\
shake\_bottle
& 97 & 13.0 & 96 & 12.1 & 96 & 9.8 & \textbf{98} & 8.8 & 96 & 8.8 & \textbf{98} & 10.1 \\
stack\_blocks\_three
& \textbf{94} & 39.8 & 74 & 37.5 & 79 & 26.9 & 85 & 23.0 & 79 & 21.4 & 82 & 25.3 \\
stack\_blocks\_two
& 97 & 27.8 & 99 & 26.3 & \textbf{100} & 18.7 & 99 & 16.6 & \textbf{100} & 15.3 & 97 & 17.9 \\
stack\_bowls\_three
& \textbf{85} & 43.8 & 76 & 43.0 & 80 & 31.2 & 79 & 27.1 & 82 & 25.6 & 82 & 31.1 \\
stack\_bowls\_two
& 97 & 30.9 & 97 & 27.9 & 96 & 20.2 & 98 & 17.2 & \textbf{99} & 16.5 & \textbf{99} & 19.6 \\
stamp\_seal
& \textbf{94} & 16.6 & 79 & 18.1 & 89 & 14.8 & 84 & 11.1 & 83 & 10.2 & 84 & 12.6 \\
turn\_switch
& \textbf{76} & 13.4 & 71 & 12.9 & 71 & 9.4 & 69 & 8.3 & 73 & 8.9 & 71 & 11.5 \\
\midrule
Average
& 89.16 & 23.5
& 85.58 & 21.21
& 88.64 & 16.0
& 88.49 & 13.9
& 88.89 & 13.3
& \textbf{89.51} & 15.7 \\
\bottomrule
\end{tabular}%
}
\end{center}
\end{table}

\end{document}

%% file: math_commands.tex
\usepackage{amsmath,amsfonts,bm}

\def\eqref#1{equation~\ref{#1}}

\def\1{\bm{1}}

\DeclareMathAlphabet{\mathsfit}{\encodingdefault}{\sfdefault}{m}{sl}
\SetMathAlphabet{\mathsfit}{bold}{\encodingdefault}{\sfdefault}{bx}{n}

